\pdfoutput=1
\documentclass[11pt]{article}

\usepackage[preprint]{acl}

\usepackage{amsmath}
\usepackage{times}
\usepackage{makecell}
\usepackage{latexsym}
\usepackage{enumitem}
\usepackage{booktabs}
\usepackage{multirow}
\usepackage{tabularx}
\usepackage{subcaption}
\usepackage[T1]{fontenc}

\usepackage[utf8]{inputenc}

\usepackage{amsmath,amssymb,amsfonts}
\usepackage{microtype}
\microtypesetup{protrusion=false} 
\usepackage{inconsolata}

\usepackage{graphicx}
\usepackage{hyperref}

\title{Repetition Mismatch: Why Data Mixture Experiments Don't Scale\\and How to Fix Them}

\author{
  Kevin Zhou\textsuperscript{$\dagger$}, Lisa Alazraki\textsuperscript{$\dagger$}, Kris Cao\textsuperscript{$\ddagger$}, Marek Rei\textsuperscript{$\dagger$} \\
    \textsuperscript{$\dagger$}Imperial College London,
    \textsuperscript{$\ddagger$}Cohere \\
    \texttt{kevinzhou497@gmail.com} \\
    \texttt{\{lisa.alazraki20, marek.rei\}@imperial.ac.uk} \\
    \texttt{kriscao@cohere.com}
}

\begin{document}
\maketitle
\begin{abstract}
Pre-training data mixtures are commonly tuned by running small-scale experiments and extrapolating to the target training budget. When high-quality data is scarce and must be repeated, this extrapolation frequently fails, but the source of the failure has not been isolated. We show that a primary culprit is a repetition mismatch: because high-quality datasets are small, their repetition rate changes as the training budget grows, shifting the optimal mixture in ways that small-scale proxy experiments do not anticipate. A subsampling procedure that matches the target repetition rate controls for this effect. In a two-source setting combining limited high-quality data with web crawl, a single repetition-controlled experiment using only 1/16 of the target tokens recovers a mixture within 0.10 of the optimum on WikiText for a 1.17B parameter model, compared to an error of 0.85 without repetition control. Achieving comparable accuracy without repetition control requires multiple training horizons, consuming 19\%, 44\%, and 94\% of the target token budget when using the results from two, three, and four horizons respectively. With three data sources, the larger mixture space requires more than a single experiment to constrain, but the approach remains effective: at the 757M scale, just two repetition-controlled horizons recover the optimal mixture, outperforming baselines that instead require the full two-source experiments to construct. Our results reveal that repetition dynamics, not scale alone, shape whether small-scale mixture experiments generalize. More broadly, they suggest that data repetition deserves treatment as a first-class variable in mixture optimization, rather than an inconvenient side effect of limited data.
\end{abstract}

\section{Introduction}

\begin{figure}[!t]
    \centering
    \includegraphics[width=0.95\linewidth]{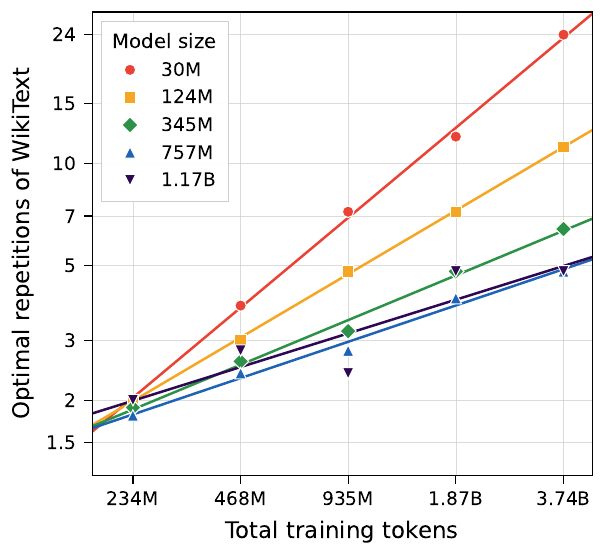}
    \caption{Optimal number of WikiText repetitions across training horizons for five model sizes. All models require similar repetition counts at small budgets, but diverge as the budget grows, causing mixtures optimized at small scale by standard extrapolation to be systematically wrong at the target scale.}
    \label{fig:wiki_rep_by_size}
\end{figure}

The composition of training data from multiple sources is a critical factor in language model (LM) pre-training, with substantial impact on downstream performance \cite{miranda2024scalediversitycoefficientdata}. Pre-training corpora typically combine noisy web crawl with cleaner, higher-quality sources such as books or curated websites, and the balance between them is a key challenge: high-quality data provides more substantive learning signals, while web crawl data helps with generalizability and regularization \cite{ICLR2024_1f7336fd, longpre-etal-2024-pretrainers}. As noted by \citet{shukor25scalinglaws}, selecting data mixtures through trial and error is costly and time-consuming, and several methods have been proposed for more efficient mixture selection \cite{xie2023, fan2024doge, ye2025mixing}. A common strategy is to run smaller-scale experiments and extrapolate the results to the target training budget, yet practitioners frequently find that mixtures tuned at small scale fail to transfer to larger regimes~\cite{kang2025autoscale}.

In this work, we identify a key factor behind this failure: \textbf{repetition mismatch}. When high-quality data is scarce, it must be repeated many times during training. Crucially, the number of repetitions changes as the training budget grows, and existing work has shown that repeated passes over a dataset can significantly affect model performance \cite{muennighoff2023scaling}. Standard scaling-based mixture selection ignores this effect: a small-scale proxy experiment imposes a fundamentally different repetition regime on the high-quality data than the target run, distorting the loss landscape and shifting the apparent optimal mixture, an effect that grows with model scale (Figure~\ref{fig:wiki_rep_by_size}). We show that controlling for this repetition mismatch largely resolves the extrapolation problem.

Our approach builds on a repetition-aware subsampling procedure first used by \citet{minimaxrepetitionaware} during pre-training. The procedure downsamples all data sources so that the high-quality data undergoes the same number of repetitions as in the full training run, while using only a fraction of the total tokens. We use this procedure to isolate repetition as a variable in mixture prediction. To test this, we compare it against a standard scaling laws-based approach that extrapolates optimal mixture ratios from shorter training runs without matching repetition rates.

Our experiments combine a limited high-quality dataset -- either WikiText~\cite{merity2016pointersentinelmixturemodels} or biomedical literature from PubMed~\cite{pubmed} -- with FineWeb~\cite{penedo24fineweb}, a large-scale web crawl corpus. We first examine the two-source case, then extend to a three-source setting using both high-quality datasets alongside FineWeb. Experiments span five model sizes (30M to 1.17B), allowing us to trace how model capacity interacts with repetition dynamics in mixture prediction. Our findings are:
\begin{itemize}
    \item \textbf{Repetition mismatch is a dominant confounder in small-scale mixture prediction.} Matching the repetition rate of the target run, rather than just reducing the training budget, is sufficient to recover accurate mixture predictions from small-scale experiments. The effect is consistent across WikiText and PubMed as high-quality sources, as well as model sizes from 124M to 1.17B. 
    \item \textbf{Repetition control enables accurate mixture prediction from minimal compute.} In the two-source setting, a single repetition-controlled experiment using $\sim$1/16 of the target horizon tokens recovers mixtures within 0.10–0.15 of the optimum for the 1.17B model across both WikiText and PubMed, compared to errors of 0.70–0.85 without repetition control. Reaching comparable accuracy without repetition control requires multiple training horizons, consuming 19\%, 44\%, and 94\% of the target token budget when using the smallest two, three, and four horizons of our experimental setup respectively.
    \item \textbf{With more data sources, the mixture space requires more experiments to constrain, but repetition control remains effective.} At 757M parameters, just two repetition-controlled horizons recover the target optimum at a fraction of the target token budget. At 124M, multiple repetition-controlled horizons outperform both baselines, with the four-horizon prediction effectively matching the optimum (loss $2.91950$ vs. $2.91820$).
    \item \textbf{Repetition rate should be an explicit knob in mixture optimization, not an incidental consequence of budget and dataset size.} Our results demonstrate that controlling for repetition dynamics, rather than treating them as a side effect of limited data, is critical for reliable mixture prediction in data-constrained regimes.
\end{itemize}

\section{Background}

\subsection{Data Mixing in Pre-training}

Pre-training corpora for language models combine multiple data sources, and the proportions assigned to each source have a substantial impact on model performance \citep{du2021, miranda2024scalediversitycoefficientdata}. Selecting effective mixtures through trial and error is costly \citep{shukor25scalinglaws}, motivating a range of methods that aim to predict good mixtures from smaller-scale experiments. These include scaling law-based approaches that fit parametric functions to predict loss under different mixture configurations \citep{ge2025bimixbivariatedatamixing, shukor25scalinglaws, ye2025mixing}, proxy model methods that learn domain weights from auxiliary training signals \citep{xie2023, fan2024doge}, and regression-based approaches that treat mixture selection as a prediction task \citep{liu2025quadmix, liu2025regmix}. Here we focus on domain-level mixing, with the goal of determining the proportion of each data source in the training mixture, rather than on strategies operating on individual samples.

\subsection{Data Repetition and Its Effects}

When high-quality data is limited, repeated passes over the same documents are often unavoidable at training time. However, this repetition has well-documented non-linear effects on learning. \citet{muennighoff2023scaling} show that for a fixed compute budget, up to approximately four repetitions of a dataset are as effective as training on new data, whereas more than four trigger diminishing returns and performance eventually plateaus. \citet{xue2023repeat} extend this analysis, finding that the severity of multi-epoch degradation depends on model size, dataset size, as well as training objective, and that larger models are more susceptible to overfitting from excessive repetition on small datasets. Standard scaling laws for pre-training \citep{hoffman2022chinchilla, kaplan2020scaling} typically assume abundant data and do not account for these repetition effects, raising questions about their applicability in data-constrained regimes.

Crucially, these findings imply that the number of times a dataset is repeated during training is not merely a side effect of a limited data budget, but a variable that actively shapes the loss landscape. When a high-quality dataset is small relative to the training budget, the repetition count changes substantially as the budget grows. This means that a small-scale proxy experiment and the target training run operate under fundamentally different repetition regimes, even when they use the same mixture proportions.

\subsection{The Repetition Mismatch Problem}
\label{sec:repetition-mismatch}

Although the effects of data repetition are well-established, existing methods for predicting optimal data mixtures from small-scale experiments do not explicitly control for these repetition dynamics. Scaling law-based approaches \citep{ge2025bimixbivariatedatamixing, shukor25scalinglaws, ye2025mixing} extrapolate performance trends across training budgets without accounting for the fact that the repetition count of constrained data sources changes between the proxy and target scales. Proxy model methods \citep{xie2023, fan2024doge} similarly learn domain weights without modeling repetition. As a result, these methods implicitly assume that the relationship between mixture proportions and performance will hold at the target scale, an assumption that breaks down when repetition dynamics differ.

\citet{minimaxrepetitionaware} introduce a repetition-aware subsampling procedure that incidentally addresses this issue: by downsampling all data sources so that the high-quality data undergoes the same number of repetitions as in the full training run, the procedure preserves repetition dynamics while using only a fraction of the total tokens. \citet{minimaxrepetitionaware} use this procedure to inform data mixture decisions during pre-training, but do not isolate repetition mismatch as a distinct phenomenon or characterize when it matters. In this work, we identify repetition mismatch as a previously under-recognized confounder in data mixing research, and show that controlling for it addresses the extrapolation failure of small-scale mixture predictions across model sizes, dataset choices, and number of data sources.

\section{Experimental Setup}

To test whether repetition mismatch explains the failure of small-scale mixture extrapolation, we conduct experiments across multiple high-quality datasets, model sizes, and numbers of training domains.\footnote{Our code is available at \url{https://github.com/kevinzhou497/data-mixing-language-models}}

\subsection{Datasets}
\label{sec:data}

We use datasets that differ in size and quality and are commonly employed in language model pre-training \cite{yang2025umoeunifyingattentionffn, bolton2024biomedlm27bparameterlanguage}, allowing us to study how repetition dynamics affect mixture prediction when combining smaller, high-quality datasets with larger, noisier sources. Additional details of each dataset are provided in Appendix \ref{sec:appendix-data}.

\paragraph{WikiText}\hspace{-6pt}\cite{merity2016pointersentinelmixturemodels} contains articles from Wikipedia's \textit{Good and Featured} list, providing a high-quality data source that contrasts with more general web crawl. In our experiments, we use the \texttt{wikitext-103-raw-v1} instance.\footnote{\url{https://huggingface.co/datasets/Salesforce/wikitext}} After tokenization, the training set contains 116,881,107 tokens. Model performance is evaluated on a held-out WikiText validation set of 131,072 tokens. 

In all experiments, we exclude web crawl data from the validation set and evaluate performance exclusively on high-quality domains. This allows us to more directly assess the effects of data repetition and mixture composition, as validation on noisy web-sourced text can obscure differences induced by mixing strategies. Focusing on curated domains provides a more stable and interpretable evaluation signal when high-quality data is the primary object of optimization, consistent with prior studies that evaluate pre-training mixtures using curated-domain validation data~\cite{muennighoff2023scaling}.

\paragraph{PubMed}\hspace{-8pt} is a collection of biomedical literature~\citep{pubmed}, with a corresponding dataset\footnote{\url{https://huggingface.co/datasets/ncbi/pubmed}} that contains citation records for its articles. Many of these records include the text of the abstract, which we use as our data samples. To roughly match the size of the WikiText training set, we sample abstracts until the total number of tokens reaches approximately 120 million, resulting in a training set of 120,000,060 tokens. Evaluation is performed on a held-out PubMed validation set of 131,072 tokens, consistent with our procedure for WikiText.

\paragraph{FineWeb}\hspace{-6pt}\cite{penedo24fineweb} is a large-scale web-crawled text corpus. \citet{penedo24fineweb} have shown that this corpus leads to stronger language model performance compared to other web crawl datasets such as RefinedWeb~\cite{penedo2023refinedweb} and C4 \cite{2019t5}.

We use the FineWeb-10BT dataset, a subsample of approximately 10 billion tokens. At this scale and with the training horizons we employ, no repetitions of the FineWeb dataset occur, creating a clear contrast with the smaller high-quality domain datasets. This reflects realistic scenarios where limited domain-specific data is supplemented by abundant web-crawled text, and where repetition is a factor for only the high-quality sources.

\subsection{Models}
\label{sec:models}

We employ a modified version of NanoGPT~\citep{Karpathy2022} from the \texttt{modded-nanogpt} repository \citep{modded_nanogpt_2024}. This architecture reproduces GPT-2, with enhancements including the Muon optimizer~\citep{jordan2024muon} and Rotary Positional Embeddings (RoPE) \cite{su2023roformerenhancedtransformerrotary}. Further details are provided in Appendix \ref{sec:appendix-model}. 

Five model sizes are considered, obtained by varying the number of layers and the embedding dimension, resulting in approximately 30 million, 124 million, 345 million, 757 million, and 1.17 billion parameter models. This range allows us to trace how model capacity interacts with repetition dynamics in mixture prediction, which is central to our analysis. Training results for different horizons are obtained separately for each model size, isolating the effect of model capacity on the severity of repetition mismatch.

\subsection{Data Mixing Objective}

Let $D = \lbrace \mathcal{D}_1, ...,\mathcal{D}_n \rbrace $ be a set of datasets and $T_\star$ be the target training horizon. The goal is to find an optimal target mixture vector \mbox{$\boldsymbol{m}^*(T_\star)=[m^*_1,...,m^*_n]$} of length $n$, with \mbox{$\sum m^*_i=1$} and \mbox{$0 \leq m^*_i \leq 1$}. 

Since model size is fixed, the task is to predict the optimal mixture at the full target horizon using the same model trained on smaller token budgets. Specifically, after obtaining optimal mixture vectors \(\{\boldsymbol{\tilde m}(T_j)\}_{j=1}^{h}\) for smaller horizons \mbox{\(0<T_1<\cdots<T_h<T_\star\)},
we aim to predict the target mixture
{\setlength{\abovedisplayskip}{20pt}\setlength{\belowdisplayskip}{20pt}
\[
\boldsymbol{m}^*(T_\star)\ \in\ \arg\min_{\boldsymbol{m}\in\Delta^{n-1}}\ \mathcal L(\boldsymbol{m};T_\star),
\]
where \mbox{\(\Delta^{n-1}=\{\boldsymbol{m}\in\mathbb{R}_{\ge 0}^n:\sum_{i=1}^n m_i=1\}\)} is the probability simplex and $\mathcal L(\boldsymbol{m};T_\star)$ is the average cross-entropy loss on held-out validation data for mixture $\boldsymbol{m}$ and horizon $T_\star$.

Our objective is to predict $\boldsymbol{m}^*(T_\star)$ as accurately as possible while minimizing the number of required experiments to reduce computational costs.

\section{Two-Source Data Mixtures}
\label{sec:two-source-experiments}

\begin{figure*}[!t]
    \centering
    \includegraphics[width=0.95\textwidth]{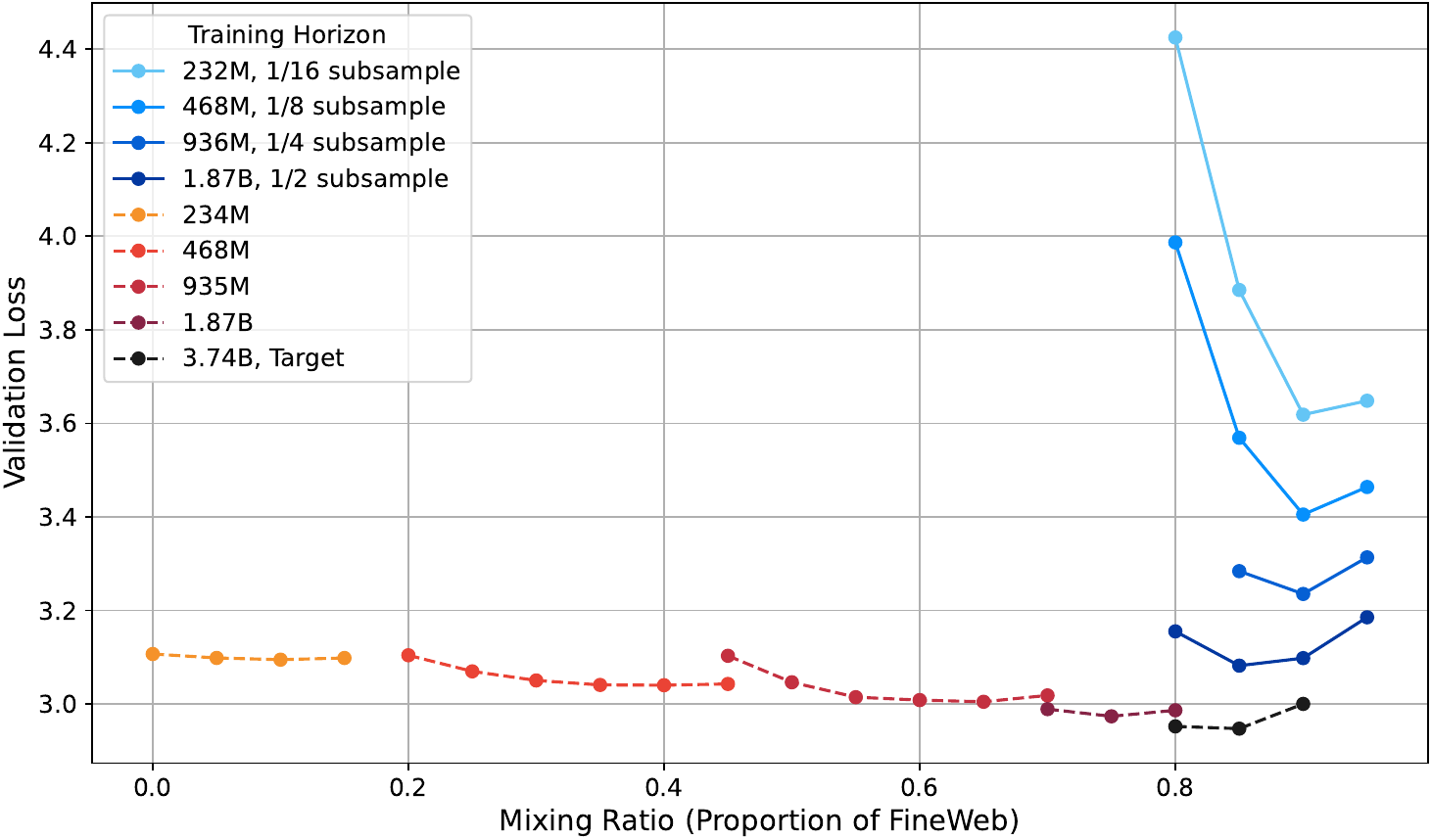}\vspace{-3pt}
    \caption{Cross-entropy loss on the validation set at the end of the training run plotted for the 757M model when using WikiText as the high-quality domain data. The scaling laws-based experiment results are shown with the dotted lines, and the repeat-aware results are shown with the solid lines. The results for other model sizes display similar behavior, and all of the optimal mixtures from our two-source experiments are presented in Appendix \ref{sec:appendix-results}.}
    \label{fig:757M_Wikipedia}
\end{figure*}

We first investigate the two-source case: high-quality data from either WikiText or PubMed combined with FineWeb. To isolate the role of repetition mismatch, we compare mixture predictions obtained with and without repetition control, holding all other experimental variables constant.

\paragraph{Without Repetition Control (Scaling Laws).}
We identify the optimal data mixture at each of 5 training horizons, with each subsequent horizon approximately doubling in length. For each horizon, we measure the optimal number of repetitions of the high-quality dataset, equivalent to the optimal mixing ratio in the two-source case. We then predict the target-horizon mixture in four ways: (i) using only the smallest horizon as a direct extrapolation; (ii--iv) fitting a linear regression over the two, three, or four smallest horizons between training tokens and optimal repetitions. Unlike prior data mixing scaling laws, which predict loss or perplexity, we predict the optimal mixing ratio directly, as our goal is to evaluate how accurately small-scale experiments recover the target mixture. This linear-extrapolation baseline formalizes the practitioner workflow of running mixture sweeps at smaller scales and projecting forward, used in industry pre-training pipelines such as Llama 3 \cite{grattafiori2024llama3herdmodels} and identified as standard practice by \citet{shukor25scalinglaws} and \citet{kang2025autoscale}. Each model size is treated as a separate set of experiments. For each horizon, we sweep over mixing ratios in increments of $0.05$, training until a U-shaped curve in the validation loss emerges, indicating that the optimal ratio has been bracketed. Across experiments, the validation loss generally decreases toward the optimum and then increases; the occasional local non-monotonicity motivates our fine-grained 0.05 sweep granularity, which we find sufficient to identify the true optimum region rather than a local minimum.

\begin{table*}[t]
\scriptsize
\centering
\resizebox{0.99\textwidth}{!}{%
\begin{tabular}{@{}c c c c c c c c c c@{}}
\toprule
\textbf{High-Quality Dataset} & \textbf{Model Size} &
\multicolumn{4}{c}{\textbf{Scaling Laws Prediction Error}} &
\multicolumn{4}{c}{\textbf{Repeat-aware Prediction Error}} \\
\cmidrule(lr){3-6} \cmidrule(lr){7-10}
& & 1-H & 2-H & 3-H & 4-H & 1-H & 2-H & 3-H & 4-H \\
\midrule
\multirow{5}{*}{WikiText} 
  & 30M  & \textbf{0.250} & \textbf{0.064} & 0.060 & 0.039 & 0.550 & 0.256 & 0.013 & 0.044 \\
  & 124M & 0.650 & \textbf{0.034} & \textbf{0.006} & \textbf{0.001} & \textbf{0.200} & 0.200 & 0.097 & 0.062 \\
  & 345M & 0.750 & 0.129 & 0.013 & 0.000 & \textbf{0.100} & 0.100 & 0.011 & 0.017 \\
  & 757M & 0.750 & 0.028 & 0.010 & 0.006 & \textbf{0.050} & 0.050 & 0.050 & 0.000 \\ 
  & 1.17B   & 0.850 & 0.090 & \textbf{0.052} & \textbf{0.014} & \textbf{0.100} & 0.100 & 0.220 & 0.109 \\
\midrule
\multirow{5}{*}{PubMed} 
  & 30M  & \textbf{0.300} & \textbf{0.114} & 0.110 & 0.059 & 0.500 & 0.315 & 0.119 & 0.079 \\
  & 124M & 0.650 & 0.259 & \textbf{0.044} & 0.032 & \textbf{0.200} & \textbf{0.200} & 0.095 & 0.061 \\
  & 345M & 0.750 & 0.129 & 0.032 & 0.012 & \textbf{0.100} & 0.100 & 0.011 & 0.016 \\
  & 757M & 0.650 & \textbf{0.011} & \textbf{0.001} & 0.029 & \textbf{0.100} & 0.100 & 0.100 & 0.050 \\
  & 1.17B   & 0.700 & 0.129 & \textbf{0.090} & 0.027 & \textbf{0.150} & 0.150 & 0.150 & 0.050 \\
\bottomrule
\end{tabular}%
}
\caption{Distances from the optimal mixture across model sizes, measured by the absolute difference in FineWeb proportion across datasets and prediction horizons. Since these experiments involve only two data sources, this difference directly reflects the mixture prediction error. The better-performing method is in \textbf{bold} when the two methods differ by at least 0.05, our mixing-ratio sweep granularity; smaller differences should be counted as ties and thus neither method is bolded. The 1-H, 2-H, etc.\ column headers refer to the 1-Horizon, 2-Horizon, etc.\ predictions.}
\label{tab:distance-optimal-all}
\end{table*}

\paragraph{With Repetition Control (Repeat-aware).}
To control for repetition mismatch, we apply the subsampling procedure described in Section~\ref{sec:repetition-mismatch}, using the same target horizon and subsamples of $\frac{1}{16}, \frac{1}{8}, \frac{1}{4}, \mbox{and } \frac{1}{2}$, with subsampling performed at the document level.

To illustrate, let the target horizon contain $T_\star$ tokens, the high-quality dataset be $D$ with length $n_D$, and the mixture proportion be $h$. In the full training setup, the number of repetitions of $D$ is $\frac{T_\star \times h}{n_D}$. In the $\frac{1}{S}$ subsample scenario, only the first $\frac{1}{S}$ fraction of documents from $D$ is used, and the training horizon is reduced to $\frac{1}{S} \times T_\star$ tokens. Keeping the same mixture proportion $h$ and assuming reasonably uniform document lengths, the number of repetitions becomes
{\setlength{\abovedisplayskip}{24pt}\setlength{\belowdisplayskip}{24pt}
\[
\frac{T_\star \times \frac{1}{S} \times h}{n_D \times \frac{1}{S}} = \frac{T_\star \times h}{n_D}.
\]
}
Thus, this setup preserves the same number of repetitions as the full scenario while using only $\frac{1}{S}$ of the total tokens. A structural consequence of this identity is that repeat-aware predictions fix the target repetition count deterministically, whereas the scaling-laws baseline leaves the repetition count to vary with the proxy horizon.
We compute the optimal mixture at each subsampled horizon and predict the target mixture using the same four formulations as above. Further experimental details, including hyperparameters, are provided in Appendix \ref{sec:appendix-procedure}.

\subsection{Two-Source Results and Discussion}
\label{sec:two-source-results}

Figure \ref{fig:757M_Wikipedia} presents results for the 757M model with WikiText as the high-quality source. Training token counts across corresponding horizons differ slightly between setups, as repeat-aware subsampling operates at the document level.

Across all model sizes and both high-quality sources, a consistent pattern emerges: without repetition control, the optimal proportion of high-quality data decreases as the training budget increases, reflecting the diminishing returns from excessive repetition documented by~\citet{muennighoff2023scaling}. PubMed and WikiText follow remarkably similar trajectories, underscoring the generalizability of these findings. Full mixture results are in Appendix \ref{sec:appendix-results}.

\paragraph{Repetition Control Stabilizes Mixture Predictions.}
The most striking feature of Figure \ref{fig:757M_Wikipedia} is the tight clustering of optimal mixing ratios across horizons under repetition control, compared to the large drift without it. Table \ref{tab:distance-optimal-all} quantifies this pattern.

\paragraph{Single-Horizon Predictions.}
The single-horizon case most directly reveals the effect of repetition mismatch. Using only the smallest horizon ($\sim\frac{1}{16}$ of the target tokens), repetition control recovers a mixture within $0.100$ of the optimum for the 1.17B model on WikiText, versus $0.850$ without it. PubMed shows the same pattern ($0.150$ versus $0.700$), as does the 757M model on both domains ($0.050$ versus $0.750$ with WikiText and $0.100$ versus $0.650$ with PubMed). With WikiText, this ``one-shot'' prediction uses only $\sim$232M tokens compared to 3.74B at the target ($\sim$241M and 3.84B with PubMed), yet recovers a near-optimal mixture at a fraction of the cost. For the 124M, 345M, 757M, and 1.17B model sizes and with both WikiText and PubMed, repetition control recovers a stronger mixture with a single horizon than without it. For the smallest model (30M), repetition control performs worse.

\paragraph{Multiple Horizons.}
As additional horizons are added, both methods converge toward comparable accuracy. Repeat-aware achieves this convergence immediately: its predictions are already near the target-horizon optimum after a single horizon and generally change little as more are added, while the linear regression baseline improves gradually as it accumulates fitting points. For the 345M and 757M models this produces effective ties across the 2-, 3-, and 4-horizon regimes, with most differences at or below the 0.05 mixing-ratio sweep granularity. At 1.17B, the linear baseline continues to improve with additional horizons and eventually overtakes repeat-aware at three or four horizons by more than the 0.05 sweep granularity; but by that point the sweep has consumed 44–94\% of the target token budget (Section \ref{sec:compute_efficiency}), whereas repeat-aware has held its single-horizon prediction throughout at $\sim$6\% of the target budget. The practical value of repetition control therefore lies in this early-locking behavior: a single cheap horizon already delivers a near-optimal mixture.

\paragraph{The Role of Model Capacity.}
The benefit of repetition control depends strongly on model capacity (Table \ref{tab:distance-optimal-all}). The single-horizon improvement with WikiText shrinks from $0.750$ at 1.17B parameters to $0.450$ at 124M, and at 30M, repetition control is outperformed by scaling-law extrapolation, identifying a lower bound on the model scale where the method applies. Figure \ref{fig:wiki_rep_by_size} reveals the underlying mechanism: the optimal repetition count at the target horizon ranges from $\sim$5 for the 1.17B and 757M models to $\sim$24 for the 30M. Because subsampling preserves the repetition count while reducing the absolute number of high-quality tokens, smaller models that require many repetitions at the target horizon end up with too few unique tokens to learn from. Repetition control therefore works best when the model is large enough to extract signal from high-quality tokens efficiently. The 30M results illustrate the failure mode at very small model scales and establish this lower bound empirically; the 124M, 345M, 757M, and 1.17B results show the method works above it.

Notably, the optimal repetition counts for the 30M and 124M models substantially exceed the $\sim$4-repetition threshold identified by \citet{muennighoff2023scaling}, beyond which diminishing returns typically begin. Rather than contradicting this finding, this suggests that the threshold is model-size-dependent. The PubMed experiments exhibit the same pattern (Figure~\ref{fig:PubMed_Optimal_Epochs}, Appendix~\ref{sec:appendix-results}).

\paragraph{Bootstrap Validation of the Optimal Mixtures}
We also compute paired bootstrap 95\% confidence intervals on the loss over the validation set for the 1.17B target-horizon mixture sweeps, using 10,000 re-samples. For PubMed, the target-horizon optimum of 0.80 FineWeb and 0.20 PubMed is significantly separated from its neighboring mixtures of (0.75 FineWeb, 0.25 PubMed) and (0.85 FineWeb, 0.15 PubMed). For WikiText, the loss landscape is comparatively flat around the optimal mixture of 0.85 FineWeb and 0.15 WikiText. The optimum is significantly better than the (0.90, 0.10) mixture but not the (0.80, 0.20) mixture; however, both are significantly better than all other mixtures in the sweep, demonstrating that the true optimal mixture lies in this near-optimal region. Both sweeps also show a mild non-monotonicity at (0.75, 0.25), where the loss is higher than at either neighboring mixture; this local perturbation motivates our fine-grained sweep granularity, which the bootstrap confirms is sufficient to identify the true optimum region rather than a local minimum. The full set of bootstrap confidence intervals for the target horizon is presented in Appendix~\ref{sec:appendix-bootstrap}.

\section{Three-Source Data Mixtures}
\label{sec:three_source}

We next test whether repetition control remains effective with a larger mixture space. We use three data sources: WikiText and PubMed as high-quality datasets, and FineWeb as the web crawl. Model evaluation averages the loss on the WikiText and PubMed validation sets. This setup reflects common pre-training configurations that combine multiple high-quality sources with general web crawl, such as the data mixture used for the first LLaMA models \cite{touvron2023llama}.

\paragraph{Experimental Procedure.}
We follow the same procedure as in the two-source case, using the same subsample proportions and a full target horizon of $\sim$3.79 billion tokens. We run experiments at both the 124M and 757M model scales, mirroring the two-source setup and allowing us to test whether the model-capacity trend from Section~\ref{sec:two-source-results} carries over to the larger mixture space. Since WikiText and PubMed differ slightly in total token count, we average the iteration counts from the corresponding two-source experiments at each horizon.

\begin{table*}[!t]
\centering
\renewcommand\arraystretch{1}
\setlength{\tabcolsep}{6pt}
\scalebox{0.865}{%
\begin{tabular}{l l c c c}
\toprule
\makecell{ \textbf{Model} \textbf{Size}} &
\makecell{ \textbf{Mixing} \textbf{Ratio}} &
\makecell{ \textbf{Learning} \textbf{Rate}} &
\makecell{ \textbf{Avg.}  \textbf{Validation }\textbf{Loss}} &
\makecell{\textbf{Experiment} \textbf{Type}} \\
\midrule
\multirow{7}{*}{124M}
  & \mbox{0.3, 0.35, 0.35}   & 0.001 & 2.94270 & Baseline 1 \\
  & \mbox{\textbf{0.45, 0.25, 0.3}} & \textbf{0.001} & \textbf{2.91820} & \textbf{Optimal Mixture} \\
  & \mbox{0.51, 0.245, 0.245} & 0.001 & 2.91950 & Four-Horizon Prediction \\
  & \mbox{0.56, 0.22, 0.22}   & 0.001 & 2.92830 & Three-Horizon Prediction \\
  & \mbox{0.57, 0.215, 0.215}   & 0.001 & 2.92965 & Two-Horizon Prediction \\
  & \mbox{0.65, 0.175, 0.175}   & 0.001 & 2.95570 & Baseline 2 \\
  & \mbox{0.75, 0.125, 0.125}   & 0.00141 & 3.01300 & Single-Horizon Prediction \\
\midrule
\multirow{4}{*}{757M}
  & \mbox{\textbf{0.65, 0.175, 0.175}}  & \textbf{0.001} & \textbf{2.7699} & \textbf{Optimal / Two-Horizon Prediction} \\
  & \mbox{0.65, 0.15, 0.20}   & 0.001 & 2.7751 & Baseline 1 \\
  & \mbox{0.825, 0.075, 0.100}   & 0.001 & 2.8337 & Baseline 2 \\
  & \mbox{0.85, 0.075, 0.075}   & 0.001 & 2.8518 & Single-Horizon Prediction \\
\bottomrule
\end{tabular}%
}
\caption{Key results from the three-source repeat-aware experiments at the full training horizon for the 124M and 757M models. Mixing ratios are shown as proportions of FineWeb, WikiText, and PubMed, with the optimal mixture per model size highlighted in \textbf{bold}.}
\label{tab:mixing_results_main}
\end{table*}

\paragraph{Baselines.}
We compare against two baselines derived from the two-source results at each model scale: (i) using the optimal proportion of each high-quality domain from its respective two-source experiment, and (ii) averaging the optimal high-quality proportions across the WikiText and PubMed two-source experiments, allocated in proportion to their two-source optima. In both cases, the remainder is assigned to FineWeb. For the 124M model, both two-source optima are 0.35, yielding Baseline 1 = $[0.30, 0.35, 0.35]$ and Baseline 2 = $[0.65, 0.175, 0.175]$. For 757M, the WikiText and PubMed optima are 0.15 and 0.20, yielding Baseline 1 = $[0.65, 0.15, 0.20]$ and Baseline 2 = $[0.825, 0.075, 0.100]$. Both baselines use target-horizon two-source optima, giving them strictly more information about the two-source structure of the problem than any small-scale extrapolation would have; comparisons against them are therefore conservative.

\subsection{Three-Source Results and Discussion}
Table \ref{tab:mixing_results_main} presents the three-source results at the target horizon for the 124M and 757M models. We evaluate mixture predictions here on validation loss rather than mixing-ratio distance, since three-source mixtures cannot be reduced to a single scalar coordinate.
Following the same conservative convention we apply throughout the paper, our conclusions rest on the pattern that holds across model sizes rather than on any individual validation loss gap and we do not rely on interpreting small loss differences as providing strict rankings.

\paragraph{One horizon is the floor; more horizons rapidly close the gap.}
With three sources, we observe the trend across both model sizes that a single horizon yields mixtures that are further from the optimum, but predictions with additional horizons do recover near-optimal mixtures. A single repetition-controlled horizon yields a mixture of $[0.75, 0.125, 0.125]$ at 124M against a true optimum of $[0.45, 0.25, 0.3]$, and $[0.85, 0.075, 0.075]$ at 757M against $[0.65, 0.175, 0.175]$. Both predictions drift toward higher FineWeb proportions, leaving a loss gap of $\sim$0.08--0.10 from the optimum, consistent with the larger mixture space requiring more than one experiment to constrain. As we show next, adding even a single additional horizon dramatically narrows this gap.

\paragraph{Two horizons suffice at larger model scale.}
At 757M, two horizons close the remaining gap to the optimum. The two-horizon repetition-controlled prediction yields a FineWeb proportion of $0.65$, giving a mixture of $[0.65, 0.175, 0.175]$ that recovers the target optimum at sweep granularity (avg.\ loss $2.7699$). It outperforms Baseline 2 ($2.8337$) and matches Baseline 1 ($2.7751$). Two short repetition-controlled runs in the three-source setting recover the optimal mixture, while the closest competing baseline requires two complete two-source sweeps as a prerequisite. This mirrors the model-capacity trend observed in the two-source experiments: as model size grows, repetition control becomes increasingly sample-efficient, and the number of horizons needed to constrain the mixture space drops.

\paragraph{Multi-horizon predictions converge to the optimum at smaller scale.}
At the 124M scale, where two horizons are not yet sufficient,
additional horizons progressively close the gap to the optimum.
Fitting linear regressions over the smallest two, three, and four horizons predicts FineWeb proportions of approximately $0.57$, $0.56$, and $0.51$ respectively. Distributing the remaining budget evenly between WikiText and PubMed and training at these predicted mixtures, all three outperform both baselines (Table~\ref{tab:mixing_results_1_appendix}). The four-horizon prediction ($[0.51, 0.245, 0.245]$) achieves a loss of $2.91950$, effectively matching the true optimum of $2.91820$, while the two- and three-horizon predictions reach competitive accuracy at substantially lower cost. Together with the 757M two-horizon result, these findings show that repetition control remains effective in the three-source setting, with the number of horizons needed shrinking as model capacity grows, mirroring the same interaction between repetition dynamics and model scale that underlies the two-source results.

\begin{figure}[t]
\centering
\includegraphics[width=\columnwidth]{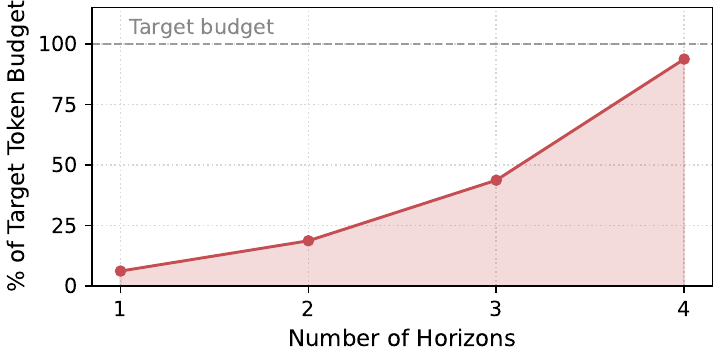}
\caption{Cumulative training tokens across horizons for the WikiText two-source experiments, as a percentage of the target token budget. Each additional horizon roughly doubles the cumulative cost: by four horizons, a sweep consumes nearly the full target budget, making the single-horizon regime the most cost-efficient when accuracy permits.}
\label{fig:compute_cost}
\end{figure}

\section{Compute Cost of Mixture Prediction}
\label{sec:compute_efficiency}

Both approaches use the same training horizons and sweep over mixing ratios and learning rates identically, so the cost at any given horizon is the same; the practical difference is how many horizons are needed. In the two-source experiments with WikiText, repetition control achieves a prediction error of just $0.05$ for the 757M model and $0.10$ for the 1.17B model using a single horizon, roughly $6\%$ of the target token budget. Without repetition control, the same horizon yields an error of $0.75$ and $0.85$ respectively. Using multiple horizons to achieve comparable accuracy requires a much greater proportion of the target budget; roughly $19\%$ for two horizons, $44\%$ for three horizons, and $94\%$ for four horizons (Figure~\ref{fig:compute_cost}). With three sources, the larger mixture space requires more than one horizon, but two repetition-controlled horizons (roughly $19\%$ of the target budget) recover the optimum at 757M and beat both baselines at 124M. The savings thus come entirely from needing fewer horizons, and grow with the target training budget; for precise per-horizon token counts and a trillion-token extrapolation, see Appendix~\ref{sec:appendix-compute}.

\section{Conclusion}

In this work, we set out to understand why data mixtures tuned at small scale often fail to transfer to larger training budgets in data-constrained settings. Our experiments point to a clear culprit: repetition mismatch. When high-quality data is scarce, it must be repeated during training, and the number of repetitions changes as the training budget grows. Small-scale proxy experiments and full-scale target runs therefore operate under different repetition regimes, and standard scaling-based approaches do not account for this.

Controlling for repetition resolves the problem. A single repetition-controlled horizon using ~1/16 of the target tokens recovers a mixture within 0.10–0.15 of the optimum at 1.17B and 0.05–0.10 at 757M across both WikiText and PubMed, compared to errors of 0.70–0.85 and 0.65–0.75 respectively without repetition control. This advantage strengthens with model capacity, and in a three-source setting, as few as two horizons match the target optimum at larger scale. Each of these results required holding repetition fixed between proxy and target experiments rather than letting it drift with the training budget.

Repetition control is also a simple intervention, operating at the dataset level and requiring no parametric modeling, proxy training runs, or hyperparameter tuning beyond the mixing-ratio sweep that any mixture prediction method already performs. It is therefore orthogonal to existing approaches and could be incorporated into any of them, yet to our knowledge no current method does so.

More broadly, these findings suggest that data repetition deserves to be treated as a primary variable in mixture optimization rather than an inconvenient side effect of limited data. Methods that predict mixtures from small-scale proxy experiments in data-constrained regimes should control for repetition dynamics, as failing to do so risks systematic prediction errors that grow with both model capacity and the gap between proxy and target scales. As practitioners increasingly rely on smaller ablation runs to inform mixture decisions for billion-parameter models, accounting for repetition mismatch will only become more important.

\section*{Limitations}
\label{sec:limitations}
Our experiments use models up to 1.17B parameters and horizons up to $\sim$3.8B tokens, smaller than modern LLM pre-training. Extrapolating trends from smaller scales is standard practice in scaling laws research \citep{kaplan2020scaling, hoffman2022chinchilla}, and repetition control's general strengthening with model size across the range we study suggests these findings extend to multi-billion parameter models. Empirical confirmation at that scale is beyond our resources, and we leave it as a direction for future investigation.

Our results are based on single runs per configuration, consistent with prevailing practice in scaling laws and data mixing research \citep{bordt2026train, pmlr-v267-magnusson25a}, where pre-training costs make multi-seed sweeps impractical at our mixture-ratio and horizon grid scales. Our claims accordingly rest on aggregate trends that hold across both high-quality datasets and the range of model sizes we study. As noted in Section~\ref{sec:two-source-results}, individual differences within the 0.05 mixing-ratio sweep granularity should be interpreted conservatively. A direct check on run-to-run variation shows that our conservative treatment is well within actual noise margins: rerunning five PubMed configurations under each method at three seeds each shifted validation loss by at most 0.0085, well below the differences underlying our main claims. Full multi-seed results are reported in Appendix~\ref{sec:appendix-seed-variance}.

We compare repetition control against a linear-regression baseline over optimal mixing ratios across horizons, formalizing the practitioner workflow standard in industry pipelines (see Section~\ref{sec:two-source-experiments}). No published mixture-prediction method, to our knowledge, includes per-source repetition as an input variable, including the parametric approaches of \citet{ye2025mixing}, \citet{shukor25scalinglaws}, \citet{ge2025bimixbivariatedatamixing}, \citet{kang2025autoscale}, and \citet{liu2025regmix}. These methods could incorporate repetition-awareness through several natural routes: as an input feature in a RegMix-style regression, or as a repetition-dependent term in a parametric data-mixing law fit jointly with the mixture coefficients. The intervention is additive to existing methods rather than a replacement, and we view it as a natural extension of our findings.

Our evaluation focuses on validation loss over the high-quality domains (WikiText and PubMed). This choice is consistent with prior data mixing work \citep{ye2025mixing, muennighoff2023scaling} and reflects a methodological consideration in our setup: because FineWeb is unrepeated across all horizons, web-crawl perplexity primarily tracks exposure to FineWeb tokens rather than mixture quality, while the signal that distinguishes mixtures is most cleanly observable on the repeated high-quality sources. Our accuracy claims are therefore scoped to this objective: a mixture identified as optimal here minimizes high-quality-domain validation loss, and whether it is optimal under downstream-task or full-corpus objectives is an empirical question we do not answer. Confirming that our findings transfer to downstream benchmarks remains an important direction for future work.

Our experiments combine one or two high-quality datasets with a single large web crawl, a simplified setup compared to real-world pre-training corpora, which typically draw from seven or more sources with varying repetition rates \citep{touvron2023llama, weber2024redpajama}. This few-source design follows prior work studying repetition effects and data composition in controlled settings \citep{muennighoff2023scaling, xue2023repeat}. A controlled few-source setup also avoids confounds like content overlap or near-duplication that can obscure repetition effects; here, each high-quality source is repeated independently and disjointly from the web crawl, isolating the repetition variable cleanly. The repetition-aware procedure generalizes naturally to any number of sources, though we leave its empirical behaviour on more diverse mixtures with datasets of varied sizes to future investigation.

All experiments use English-language datasets. Since data scarcity is often more acute for non-English languages~\cite{joshi-etal-2020-state}, repetition mismatch may be an even greater concern in multilingual settings, which we leave to future work.

\section*{Ethical Considerations}
All datasets employed in our experiments are publicly available and have been used in accordance with their respective licenses. Our models are based on open-source architectures and are trained using open-source software released under permissive licenses. Additionally, we share our code publicly to support reproducibility.

Pre-training language models is computationally expensive, and the experimental sweeps required to study data mixing compound this cost. Our results show that controlling for repetition can reduce the experimental budget needed to identify effective mixtures. This may help reduce the computational and environmental cost of mixture selection as data mixing studies become standard practice for billion-parameter pre-training.

Our PubMed experiments use abstracts from the publicly released HuggingFace PubMed dataset. We did not extract or process any personally identifiable information, and all biomedical content used is publicly distributed for research purposes.

\bibliography{custom}

\clearpage
\appendix

\section{Dataset Details}
\label{sec:appendix-data}

All datasets are tokenized with tiktoken\footnote{\texttt{\url{https://github.com/openai/tiktoken}}} using GPT-2 encodings.

\subsection{WikiText}
WikiText consists of full-length articles, making it well-suited for evaluating models on long-range dependencies. Performance on the validation set thus partly reflects the model's ability to capture relationships across longer spans within a document.

\subsection{PubMed}
We select PubMed as a second high-quality domain for two reasons. First, the text comes from published academic articles that undergo rigorous review, so the samples are consistently well-written, comparable in quality to the \textit{Good and Featured} WikiText articles. Second, PubMed is domain-specific: biomedical literature contains specialized terminology in physiology, medicine, and related fields that is rare in general discourse. This combination of high quality and domain specificity makes PubMed a useful complement to WikiText.

The Hugging Face PubMed dataset does not provide a predefined train/validation split, so we hold out approximately $1\%$ of documents for validation. The full corpus contains 6,435,414,914 training tokens and 65,039,475 validation tokens. To maintain comparable dataset sizes with WikiText, we sample abstracts until the training set reaches approximately 120 million tokens (120,000,060), and construct a validation set of 200,191 tokens. As noted in the main paper, the number of validation tokens used in evaluation is fixed at 131,072.

\subsection{FineWeb}
FineWeb is a large-scale dataset of 18.5 trillion tokens of cleaned and deduplicated web crawl data from Common Crawl\footnote{\texttt{\url{https://commoncrawl.org}}}. We use the FineWeb-10BT subset, a random sample of approximately 10 billion tokens, as described in Section \ref{sec:data}.

We chose FineWeb-10BT over alternatives such as FineWeb-Edu \cite{penedo24fineweb} because the web crawl data in our experiments serves primarily as a source of regularization and generalizability. The more general FineWeb-10BT corpus aligns with this role and maintains a clear quality contrast with WikiText and PubMed.

\section{Model Details}
\label{sec:appendix-model}

\subsection{NanoGPT}
NanoGPT \cite{Karpathy2022} provides a streamlined framework for training medium-sized GPT models. The \texttt{modded-nanogpt} repository \cite{modded_nanogpt_2024} hosts a speedrun challenge where practitioners train a language model to reach a target loss on FineWeb as quickly as possible.

We use a version that adds two modifications to the base GPT-2 architecture: the Muon optimizer \cite{jordan2024muon} and Rotary Positional Embeddings (RoPE) \cite{su2023roformerenhancedtransformerrotary}. Muon (MomentUm Orthogonalized by Newton-Schulz) applies a Newton-Schulz matrix iteration \cite{bernstein2024newtonschulz} to SGD-momentum updates. In our setup, Muon optimizes the two-dimensional weight matrices in the hidden layers, while AdamW handles the remaining parameters (embedding layer, final fully connected layer). RoPE encodes relative positions by rotating query and key vectors, improving training efficiency and robustness.

These modifications correspond to the October 11, 2024 pull request in the \texttt{modded-nanogpt} repository.\footnote{\url{https://github.com/KellerJordan/modded-nanogpt/commit/b356a1f}}

\subsection{Model Sizes}
We use five model sizes, obtained by scaling the number of layers and embedding dimensions. The 124M model follows the original \texttt{modded-nanogpt} configuration ($12$ layers, $768$-dimensional embeddings, 123,532,032 parameters). The 30M model scales both down by $50\%$ ($6$ layers, $384$ dimensions, 29,915,520 parameters). The 345M model scales both up by $50\%$ ($18$ layers, $1152$ dimensions, 344,550,528 parameters). The 757M model uses $24$ layers and $1536$-dimensional embeddings (756,672,000 parameters). The 1.17B model uses $28$ layers and $1792$-dimensional embeddings (1,169,045,248 parameters).

\begin{table}[!t]
\centering
\scalebox{0.78}{%
\begin{tabular}{@{}c c c c@{}}
\toprule
\makecell{\textbf{High-Quality}\\\textbf{Dataset}} & \makecell{\textbf{Model}\\\textbf{Size}} & \makecell{\textbf{Training}\\\textbf{Tokens}} & \makecell{\textbf{Optimal}\\\textbf{Mixing Ratio}} \\
\midrule
\multirow{25}{*}{WikiText}
  & \multirow{5}{*}{30M}
    & 234M            & (0.00, 1.00) \\
  & & 468M            & (0.05, 0.95) \\
  & & 935M            & (0.10, 0.90) \\
  & & 1.87B           & (0.25, 0.75) \\
  & & 3.74B (Target)  & (0.25, 0.75) \\
\cmidrule(l){2-4}
  & \multirow{5}{*}{124M}
    & 234M            & (0.00, 1.00) \\
  & & 468M            & (0.25, 0.75) \\
  & & 935M            & (0.40, 0.60) \\
  & & 1.87B           & (0.55, 0.45) \\
  & & 3.74B (Target)  & (0.65, 0.35) \\
\cmidrule(l){2-4}
  & \multirow{5}{*}{345M}
    & 234M            & (0.05, 0.95) \\
  & & 468M            & (0.35, 0.65) \\
  & & 935M            & (0.60, 0.40) \\
  & & 1.87B           & (0.70, 0.30) \\
  & & 3.74B (Target)  & (0.80, 0.20) \\
\cmidrule(l){2-4}
  & \multirow{5}{*}{757M}
    & 234M            & (0.10, 0.90) \\
  & & 468M            & (0.40, 0.60) \\
  & & 935M            & (0.65, 0.35) \\
  & & 1.87B           & (0.75, 0.25) \\
  & & 3.74B (Target)  & (0.85, 0.15) \\
\cmidrule(l){2-4}
  & \multirow{5}{*}{1.17B}
    & 234M            & (0.00, 1.00) \\
  & & 468M            & (0.30, 0.70) \\
  & & 935M            & (0.70, 0.30) \\
  & & 1.87B           & (0.70, 0.30) \\
  & & 3.74B (Target)  & (0.85, 0.15) \\
\midrule
\multirow{25}{*}{PubMed}
  & \multirow{5}{*}{30M}
    & 240M            & (0.00, 1.00) \\
  & & 480M            & (0.05, 0.95) \\
  & & 960M            & (0.10, 0.90) \\
  & & 1.92B           & (0.20, 0.80) \\
  & & 3.84B (Target)  & (0.30, 0.70) \\
\cmidrule(l){2-4}
  & \multirow{5}{*}{124M}
    & 240M            & (0.00, 1.00) \\
  & & 480M            & (0.15, 0.85) \\
  & & 960M            & (0.40, 0.60) \\
  & & 1.92B           & (0.55, 0.45) \\
  & & 3.84B (Target)  & (0.65, 0.35) \\
\cmidrule(l){2-4}
  & \multirow{5}{*}{345M}
    & 240M            & (0.05, 0.95) \\
  & & 480M            & (0.30, 0.70) \\
  & & 960M            & (0.55, 0.45) \\
  & & 1.92B           & (0.70, 0.30) \\
  & & 3.84B (Target)  & (0.80, 0.20) \\
\cmidrule(l){2-4}
  & \multirow{5}{*}{757M}
    & 240M            & (0.15, 0.85) \\
  & & 480M            & (0.40, 0.60) \\
  & & 960M            & (0.60, 0.40) \\
  & & 1.92B           & (0.75, 0.25) \\
  & & 3.84B (Target)  & (0.80, 0.20) \\
\cmidrule(l){2-4}
  & \multirow{5}{*}{1.17B}
    & 240M            & (0.10, 0.90) \\
  & & 480M            & (0.30, 0.70) \\
  & & 960M            & (0.70, 0.30) \\
  & & 1.92B           & (0.70, 0.30) \\
  & & 3.84B (Target)  & (0.80, 0.20) \\
\bottomrule
\end{tabular}%
}
\caption{Optimal mixing ratios by high-quality dataset, model size, and training token budget for the two-source scaling laws experiments.}
\label{tab:two-source-scaling}
\end{table}

\begin{table}[!t]
\centering
\scalebox{0.78}{%
\begin{tabular}{@{}c c c c@{}}
\toprule
\makecell{\textbf{High-Quality}\\\textbf{Dataset}} & \makecell{\textbf{Model}\\\textbf{Size}} & \makecell{\textbf{Training}\\\textbf{Tokens}} & \makecell{\textbf{Optimal}\\\textbf{Mixing Ratio}} \\
\midrule
\multirow{25}{*}{WikiText}
  & \multirow{5}{*}{30M}
    & 234M            & (0.80, 0.20) \\
  & & 468M            & (0.75, 0.25) \\
  & & 935M            & (0.60, 0.40) \\
  & & 1.87B           & (0.50, 0.50) \\
  & & 3.74B (Target)  & (0.25, 0.75) \\
\cmidrule(l){2-4}
  & \multirow{5}{*}{124M}
    & 234M            & (0.85, 0.15) \\
  & & 468M            & (0.85, 0.15) \\
  & & 935M            & (0.80, 0.20) \\
  & & 1.87B           & (0.75, 0.25) \\
  & & 3.74B (Target)  & (0.65, 0.35) \\
\cmidrule(l){2-4}
  & \multirow{5}{*}{345M}
    & 234M            & (0.90, 0.10) \\
  & & 468M            & (0.90, 0.10) \\
  & & 935M            & (0.85, 0.15) \\
  & & 1.87B           & (0.85, 0.15) \\
  & & 3.74B (Target)  & (0.80, 0.20) \\
\cmidrule(l){2-4}
  & \multirow{5}{*}{757M}
    & 234M            & (0.90, 0.10) \\
  & & 468M            & (0.90, 0.10) \\
  & & 935M            & (0.90, 0.10) \\
  & & 1.87B           & (0.85, 0.15) \\
  & & 3.74B (Target)  & (0.80, 0.20) \\
\cmidrule(l){2-4}
  & \multirow{5}{*}{1.17B}
    & 234M            & (0.95, 0.05) \\
  & & 468M            & (0.95, 0.05) \\
  & & 935M            & (0.85, 0.15) \\
  & & 1.87B           & (0.85, 0.15) \\
  & & 3.74B (Target)  & (0.85, 0.15) \\
\midrule
\multirow{25}{*}{PubMed}
  & \multirow{5}{*}{30M}
    & 241M            & (0.80, 0.20) \\
  & & 481M            & (0.70, 0.30) \\
  & & 958M            & (0.55, 0.45) \\
  & & 1.92B           & (0.45, 0.55) \\
  & & 3.84B (Target)  & (0.30, 0.70) \\
\cmidrule(l){2-4}
  & \multirow{5}{*}{124M}
    & 241M            & (0.85, 0.15) \\
  & & 481M            & (0.85, 0.15) \\
  & & 958M            & (0.80, 0.20) \\
  & & 1.92B           & (0.75, 0.25) \\
  & & 3.84B (Target)  & (0.65, 0.35) \\
\cmidrule(l){2-4}
  & \multirow{5}{*}{345M}
    & 241M            & (0.90, 0.10) \\
  & & 481M            & (0.90, 0.10) \\
  & & 958M            & (0.85, 0.15) \\
  & & 1.92B           & (0.85, 0.15) \\
  & & 3.84B (Target)  & (0.80, 0.20) \\
\cmidrule(l){2-4}
  & \multirow{5}{*}{757M}
    & 241M            & (0.90, 0.10) \\
  & & 481M            & (0.90, 0.10) \\
  & & 958M            & (0.90, 0.10) \\
  & & 1.92B           & (0.85, 0.15) \\
  & & 3.84B (Target)  & (0.80, 0.20) \\
\cmidrule(l){2-4}
  & \multirow{5}{*}{1.17B}
    & 241M            & (0.95, 0.05) \\
  & & 481M            & (0.95, 0.05) \\
  & & 958M            & (0.95, 0.05) \\
  & & 1.92B           & (0.85, 0.15) \\
  & & 3.84B (Target)  & (0.80, 0.20) \\
\bottomrule
\end{tabular}%
}
\caption{Optimal mixing ratios by high-quality dataset, model size, and training token budget for the two-source repeat-aware experiments.} 
\label{tab:two-source-repeat-aware}
\end{table}

\section{Training Details}
\label{sec:appendix-procedure}

\subsection{Hyperparameters}
We use a batch size of $128$ and a sequence length of $256$ to fit within GPU memory constraints.
For the learning rate schedule, we apply a linear decay over the final $\frac{1800}{6200} \times \text{(number of iterations)}$ steps, following the \texttt{modded-nanogpt} convention. For sweeps, we typically train with three learning rates, evenly and logarithmically spaced. Exceptions occur for the 757M and 1.17B models and the longest horizons of the 124M and 345M models, where only a single learning rate is used due to resource constraints. Subsequent sweep ranges are informed by previous results; for example, if $0.00141$ is optimal in $[0.00141, 0.002, 0.00282]$, the next horizon may use $[0.001, 0.00141, 0.002]$, reflecting the trend that optimal learning rates decrease for longer horizons. Across our sweeps, the optimal learning rate at a given horizon was largely stable across mixing ratios, indicating that the mixing-ratio and learning-rate axes are largely decoupled.

\subsection{Compute Resources}
Experiments on the 30M, 124M, 345M and 757M models were conducted on NVIDIA A100 GPUs (SXM4, 80GB). The 1.17B experiments were run across two compute clusters, on NVIDIA A100 (SXM4, 80GB) and NVIDIA H200 (SXM, 141GB) GPUs respectively.

Every 1.17B run uses two data-parallel workers with a per-device microbatch of 32 sequences, independently of the GPUs available on each cluster, with gradient accumulation making up the batch size given above. We hold the number of workers fixed because our data loader assigns each worker a strided slice of the token stream, with both the per-worker offset and the stride determined by the world size: changing the number of devices changes which sequences are grouped into each batch even with the seed, the data and the batch size held fixed. Holding the world size constant keeps the sequence of training batches identical across clusters; remaining differences are limited to floating-point non-determinism in the underlying kernels.

The random seed is fixed at 42 for both data pre-processing and model training.

\subsection{Mixing Ratios}
The mixing ratio specifies the proportion of training tokens from each data source, enforced at the batch level.

For each horizon, we sweep over mixing ratios in increments of $0.05$ until the optimal ratio is identified. We occasionally deviate from $0.05$ increments when an exploratory run at a coarser spacing clearly outperforms the previous ratio, in which case intermediate values would not meaningfully affect the search. We stop the sweep once a U-shaped curve in validation loss emerges, defined as a ratio that outperforms the ratios on either side. As mentioned in Section~\ref{sec:two-source-experiments}, across experiments, the validation loss generally decreases until the optimum and then increases. This behavior signals that this is a reliable stopping criterion. 

\begin{figure}[!t]
    \centering
    \includegraphics[width=0.9\columnwidth]{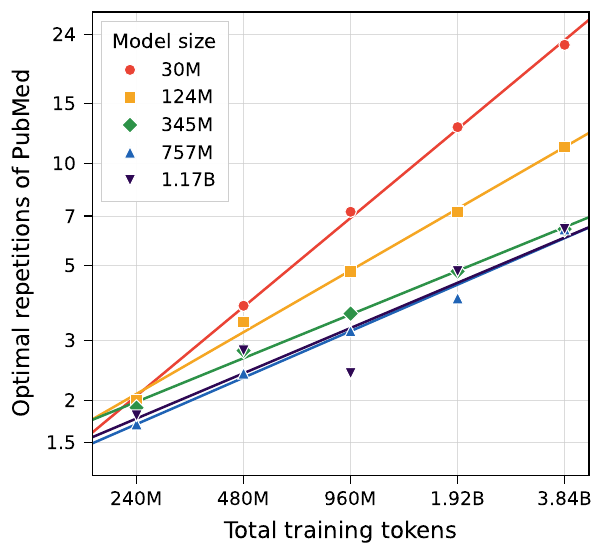}
    \caption{Optimal number of PubMed repetitions across training horizons for the 30M, 124M, 345M, 757M and 1.17B model experiments.}
    \label{fig:PubMed_Optimal_Epochs}
\end{figure}

\section{Additional Results}
\label{sec:appendix-results}

\subsection{Two-Source Mixtures}
Tables \ref{tab:two-source-scaling} and \ref{tab:two-source-repeat-aware} present the optimal mixing ratios for all scaling laws and repeat-aware experiments, respectively. Figure \ref{fig:PubMed_Optimal_Epochs} shows the optimal number of PubMed repetitions across model sizes, as discussed in Section \ref{sec:two-source-results}.

\subsection{Three-Source Mixtures}
Mixing ratios are reported as proportions of FineWeb, WikiText, and PubMed, with the best result across learning rates for each ratio. The ``Experiment Type'' column indicates whether the ratio corresponds to a baseline, the tuning procedure, or a multi-horizon prediction. For the 124M model, Tables~\ref{tab:mixing_results_16}, \ref{tab:mixing_results_8}, \ref{tab:mixing_results_4}, and \ref{tab:mixing_results_2} present results for the $\frac{1}{16}$, $\frac{1}{8}$, $\frac{1}{4}$, and $\frac{1}{2}$ subsamples; Tables~\ref{tab:mixing_results_16_757M}, \ref{tab:mixing_results_8_757M}, \ref{tab:mixing_results_4_757M}, and \ref{tab:mixing_results_2_757M} present the corresponding results for the 757M model. Target-horizon results for both models are combined in Table~\ref{tab:mixing_results_1_appendix}.

\section{Compute Cost Details}
\label{sec:appendix-compute}

Table~\ref{tab:compute_cost} reports cumulative training tokens across horizons for the WikiText two-source experiments in absolute terms. Each horizon involves multiple training runs (one per mixing ratio and learning rate), so the total experimental cost at a given horizon is the token budget shown here multiplied by the number of runs. Since both methods use the same sweep procedure, this multiplier is the same for both, and the relative cost comparison holds.

\begin{table}[!t]
\centering
\small
\begin{tabular}{lcc}
\toprule
\textbf{Horizons Used} & \textbf{Tokens per Run} & \textbf{\% of Target} \\
\midrule
1 & 232M & 6.2\% \\
2 & 700M & 18.7\% \\
3 & 1.64B & 43.7\% \\
4 & 3.50B & 93.7\% \\
\midrule
Target & 3.74B & 100\% \\
\bottomrule
\end{tabular}
\caption{Cumulative training tokens across horizons for the WikiText two-source experiments.}
\label{tab:compute_cost}
\end{table}

\begin{table}[!t]
\centering
\small
\setlength{\tabcolsep}{5pt}
\begin{subtable}{\linewidth}
\centering
\caption{Scaling laws}
\begin{tabular}{lccccc}
\toprule
\textbf{Seed} & \textbf{240M} & \textbf{480M} & \textbf{960M} & \textbf{960M} & \textbf{1.92B} \\
 & \textbf{@0.10} & \textbf{@0.30} & \textbf{@0.65} & \textbf{@0.70} & \textbf{@0.70} \\
\midrule
42 & 2.6344 & 2.6872 & 2.6720 & 2.6616 & 2.6536 \\
43 & 2.6362 & 2.6856 & 2.6745 & 2.6623 & 2.6554 \\
44 & 2.6379 & 2.6884 & 2.6745 & 2.6647 & 2.6541 \\
\midrule
\textbf{Range}  & 0.0035 & 0.0027 & 0.0025 & 0.0031 & 0.0018 \\
\textbf{SD}     & 0.0017 & 0.0014 & 0.0014 & 0.0016 & 0.0009 \\
\bottomrule
\end{tabular}
\label{tab:seed-variance-sl}
\end{subtable}
\vspace{0.5em}
\begin{subtable}{\linewidth}
\centering
\caption{Repeat-aware}
\begin{tabular}{lccccc}
\toprule
\textbf{Seed} & \textbf{241M} & \textbf{481M} & \textbf{958M} & \textbf{958M} & \textbf{1.92B} \\
 & \textbf{@0.85} & \textbf{@0.85} & \textbf{@0.80} & \textbf{@0.85} & \textbf{@0.85} \\
\midrule
42 & 3.9078 & 3.6050 & 3.3984 & 2.8991 & 2.6890 \\
43 & 3.9089 & 3.6054 & 3.4032 & 2.8972 & 2.6942 \\
44 & 3.9093 & 3.5969 & 3.3964 & 2.8994 & 2.6896 \\
\midrule
\textbf{Range}   & 0.0015 & 0.0085 & 0.0068 & 0.0021 & 0.0053 \\
\textbf{SD}      & 0.0008 & 0.0048 & 0.0035 & 0.0012 & 0.0029 \\
\bottomrule
\end{tabular}
\label{tab:seed-variance-ra}
\end{subtable}
\caption{Validation loss across three seeds (42, 43, 44) at 1.17B for PubMed sweeps, under scaling laws (a) and repeat-aware (b). Column headers give the token budget and mixing ratio (proportion of FineWeb). Scaling-laws entries use the optimal mixing ratio at each token budget, together with 960M @ 0.65 as the runner-up at that budget. Repeat-aware entries use mixing ratio 0.85 throughout, together with 958M @ 0.80 as the runner-up at that budget. Seed-to-seed variation is at most 0.0035 for scaling laws and 0.0085 for repeat-aware, in both cases well below the differences underlying our main claims.}
\label{tab:seed-variance}
\end{table}

\paragraph{Two-source setting.}
A single repetition-controlled horizon at $6\%$ of the target budget can replace a multi-horizon scaling laws analysis consuming $44$ to $94\%$ of it, since both methods perform the same number of runs per horizon. This advantage would scale with the target training budget if the per-horizon proportions held: at a 1 trillion token target, the same 6\% vs 94\% split would correspond to roughly 62 billion tokens per run for a single repetition-controlled horizon, against around 940 billion for a four-horizon scaling laws analysis. Whether these proportions hold at this scale is an empirical question beyond the range of our experiments.

\paragraph{Three-source setting.}
At the 757M scale, two repetition-controlled horizons (roughly $19\%$ of the target budget) recover the optimal mixture at sweep granularity, matching or outperforming baselines whose construction requires the full two-source experiments. At the 124M scale, repetition-controlled predictions from two to four horizons all outperform both baselines from two-source results, with the four-horizon prediction effectively reaching the optimum (loss $2.91950$ vs. $2.91820$), though at substantially higher cost. Even at this smaller scale, two horizons suffice to beat both baselines, suggesting that repetition control still substantially reduces the experimental budget needed when more data sources are involved.

\begin{table*}[!t]
\centering
\renewcommand\arraystretch{1.025}
\setlength{\tabcolsep}{25pt}
\scalebox{0.85}{%
\begin{tabular}{l c c c}
\toprule
\makecell{ \textbf{Mixing} \textbf{Ratio}} &
\makecell{ \textbf{Learning} \textbf{Rate}} &
\makecell{ \textbf{Avg.} \textbf{Validation} \textbf{Loss}} &
\makecell{\textbf{Experiment} \textbf{Type}} \\
\midrule
\mbox{0.7, 0.15, 0.15}   & 0.00141 & 3.52295 & Baseline 1 \\
\mbox{\textbf{0.75, 0.125, 0.125}} & \textbf{0.00141} & \textbf{3.50460} & \textbf{Tuned} \\
\mbox{0.75, 0.1, 0.15}     & 0.00141 & 3.51945 & Tuned \\
\mbox{0.75, 0.15, 0.1}& 0.00141 & 3.51950 & Tuned \\
\mbox{0.8, 0.1, 0.1}& 0.00141 & 3.51725 & Tuned \\
\mbox{0.85, 0.075, 0.075}& 0.002 & 3.54890 & Baseline 2 \\
\mbox{0.9, 0.05, 0.05}& 0.00141 & 3.62840 & Tuned \\
\bottomrule
\end{tabular}%
}
\caption{Three-source repeat-aware results with a $1/16$ subsample for the 124M model. Mixing ratios are proportions of FineWeb, WikiText, and PubMed. Best run in \textbf{bold}.}
\label{tab:mixing_results_16}
\end{table*}

\begin{table*}[!t]
\centering
\renewcommand\arraystretch{1.025}
\setlength{\tabcolsep}{25pt}
\scalebox{0.85}{%
\begin{tabular}{l c c c}
\toprule
\makecell{ \textbf{Mixing} \textbf{Ratio}} &
\makecell{ \textbf{Learning} \textbf{Rate}} &
\makecell{ \textbf{Avg.} \textbf{Validation} \textbf{Loss}} &
\makecell{\textbf{Experiment} \textbf{Type}} \\
\midrule
\mbox{0.45, 0.275, 0.275}   & 0.00141 & 3.66295 & Tuned \\
\mbox{0.5, 0.25, 0.25}   & 0.00141 & 3.53470 & Tuned \\
\mbox{0.55, 0.225, 0.275}   & 0.00141 & 3.44795 & Tuned \\
\mbox{0.6, 0.2, 0.2}   & 0.00141 & 3.38440 & Tuned \\
\mbox{0.65, 0.175, 0.175}   & 0.00141 & 3.34305 & Tuned \\
\mbox{\textbf{0.7, 0.15, 0.15}}  & \textbf{0.00141} & \textbf{3.32235} & \textbf{Baseline 1} \\
\mbox{0.7, 0.2, 0.1}     & 0.00141 & 3.35140 & Tuned \\
\mbox{0.7, 0.1, 0.2}& 0.00141 & 3.37140 & Tuned \\
\mbox{0.75, 0.125, 0.125}& 0.00141 & 3.33015 & Tuned \\
\mbox{0.85, 0.075, 0.075}& 0.00141 & 3.38965 & Baseline 2 \\
\bottomrule
\end{tabular}%
}
\caption{Three-source repeat-aware results with a $1/8$ subsample for the 124M model. Mixing ratios are proportions of FineWeb, WikiText, and PubMed. Best run in \textbf{bold}.} 
\label{tab:mixing_results_8}
\end{table*}

\begin{table*}[!t]
\centering
\renewcommand\arraystretch{1.025}
\setlength{\tabcolsep}{25pt}
\scalebox{0.85}{%
\begin{tabular}{l c c c}
\toprule
\makecell{ \textbf{Mixing} \textbf{Ratio}} &
\makecell{ \textbf{Learning} \textbf{Rate}} &
\makecell{ \textbf{Avg.} \textbf{Validation} \textbf{Loss}} &
\makecell{\textbf{Experiment} \textbf{Type}} \\
\midrule
\mbox{0.5, 0.25, 0.25}   & 0.00141 & 3.23315 & Tuned \\
\mbox{0.6, 0.2, 0.2}     & 0.00141 & 3.17525 & Baseline 1 \\
\mbox{\textbf{0.65, 0.175, 0.175}}& \textbf{0.00141} & \textbf{3.16845} & \textbf{Tuned} \\
\mbox{0.7, 0.15, 0.15}& 0.00141 & 3.16900 & Tuned \\
\mbox{0.7, 0.2, 0.1}& 0.00141 & 3.18480 & Tuned \\
\mbox{0.7, 0.1, 0.2}& 0.00141 & 3.20185 & Tuned \\
\mbox{0.75, 0.125, 0.125}& 0.00141 & 3.19185 & Tuned \\
\mbox{0.8, 0.1, 0.1}& 0.00141 & 3.21795 & Baseline 2 \\
\bottomrule
\end{tabular}%
}
\caption{Three-source repeat-aware results with a $1/4$ subsample for the 124M model. Mixing ratios are proportions of FineWeb, WikiText, and PubMed. Best run in \textbf{bold}.} 
\label{tab:mixing_results_4}
\end{table*}

\begin{table*}[!t]
\centering
\renewcommand\arraystretch{1.025}
\setlength{\tabcolsep}{25pt}
\scalebox{0.85}{%
\begin{tabular}{l c c c}
\toprule
\makecell{ \textbf{Mixing} \textbf{Ratio}} &
\makecell{ \textbf{Learning} \textbf{Rate}} &
\makecell{ \textbf{Avg.} \textbf{Validation} \textbf{Loss}} &
\makecell{\textbf{Experiment} \textbf{Type}} \\
\midrule
\mbox{0.45, 0.275, 0.275}   & 0.001 & 3.05655 & Tuned \\
\mbox{0.5, 0.25, 0.25}   & 0.001 & 3.03700 & Baseline 1 \\
\mbox{\textbf{0.55, 0.225, 0.225}}   & \textbf{0.001} & \textbf{3.03345} & \textbf{Tuned} \\
\mbox{0.55, 0.275, 0.175}   & 0.001 & 3.04290 & Tuned \\
\mbox{0.55, 0.175, 0.275}   & 0.001 & 3.05045 & Tuned \\
\mbox{0.6, 0.2, 0.2}   & 0.001 & 3.03565 & Tuned \\
\mbox{0.75, 0.125, 0.125}   & 0.001 & 3.08405 & Baseline 2 \\
\bottomrule
\end{tabular}%
}
\caption{Three-source repeat-aware results with a $1/2$ subsample for the 124M model. Mixing ratios are proportions of FineWeb, WikiText, and PubMed. Best run in \textbf{bold}.} 
\label{tab:mixing_results_2}
\end{table*}

\begin{table*}[!t]
\centering
\renewcommand\arraystretch{1.025}
\setlength{\tabcolsep}{25pt}
\scalebox{0.85}{%
\begin{tabular}{l c c c}
\toprule
\makecell{ \textbf{Mixing} \textbf{Ratio}} &
\makecell{ \textbf{Learning}\\\textbf{Rate}} &
\makecell{ \textbf{Average}\\  \textbf{Validation }\textbf{Loss}} &
\makecell{\textbf{Experiment}\\\textbf{Type}} \\
\midrule
\multicolumn{4}{c}{\textbf{124M Model}} \\
\midrule
\mbox{0.3, 0.35, 0.35}   & 0.001 & 2.94270 & Baseline 1 \\
\mbox{0.4, 0.3, 0.3}   & 0.001 & 2.92300 & Tuned \\
\mbox{\textbf{0.45, 0.25, 0.3}} & \textbf{0.001} & \textbf{2.91820} & \textbf{Tuned} \\
\mbox{0.45, 0.3, 0.25}   & 0.001 & 2.91915 & Tuned \\
\mbox{0.45, 0.275, 0.275}   & 0.001 & 2.91935 & Tuned \\
\mbox{0.5, 0.25, 0.25}   & 0.001 & 2.92115 & Tuned \\
\mbox{0.51, 0.245, 0.245} & 0.001 & 2.91950 & Four-Horizon Prediction \\ 
\mbox{0.56, 0.22, 0.22}   & 0.001 & 2.92830 & Three-Horizon Prediction \\
\mbox{0.57, 0.215, 0.215}   & 0.001 & 2.92965 & Two-Horizon Prediction \\
\mbox{0.6, 0.2, 0.2}   & 0.00141 & 2.94150 & Tuned \\
\mbox{0.65, 0.175, 0.175}   & 0.001 & 2.95570 & Baseline 2 \\
\mbox{0.75, 0.125, 0.125}   & 0.00141 & 3.01300 & Single-Horizon Prediction \\
\midrule
\multicolumn{4}{c}{\textbf{757M Model}} \\
\midrule
\mbox{0.55, 0.225, 0.225}   & 0.001 & 2.81550 & Tuned \\
\mbox{0.6, 0.2, 0.2}   & 0.001 & 2.78805 & Tuned \\
\mbox{\textbf{0.65, 0.175, 0.175}}   & \textbf{0.001} & \textbf{2.76990} & \textbf{Two-Horizon Prediction} \\
\mbox{0.65, 0.15, 0.2}   & 0.001 & 2.77510 & Baseline 1 \\
\mbox{0.7, 0.15, 0.15}   & 0.001 & 2.77015 & Tuned \\
\mbox{0.7, 0.2, 0.1}   & 0.001 & 2.78765 & Tuned \\
\mbox{0.7, 0.1, 0.2}   & 0.001 & 2.80405 & Tuned \\
\mbox{0.75, 0.125, 0.125}   & 0.001 & 2.78580 & Tuned \\
\mbox{0.825, 0.075, 0.1}   & 0.001 & 2.83365 & Baseline 2 \\
\mbox{0.85, 0.075, 0.075}   & 0.001 & 2.85175 & Single-Horizon Prediction \\
\bottomrule
\end{tabular}%
}
\caption{Three-source results at the full training horizon for the 124M and 757M models. Mixing ratios are proportions of FineWeb, WikiText, and PubMed. Best run per model in \textbf{bold}.} 
\label{tab:mixing_results_1_appendix}
\end{table*}

\begin{table*}[!t]
\centering
\renewcommand\arraystretch{1.025}
\setlength{\tabcolsep}{25pt}
\scalebox{0.85}{%
\begin{tabular}{l c c c}
\toprule
\makecell{ \textbf{Mixing} \textbf{Ratio}} &
\makecell{ \textbf{Learning} \textbf{Rate}} &
\makecell{ \textbf{Avg.} \textbf{Validation} \textbf{Loss}} &
\makecell{\textbf{Experiment} \textbf{Type}} \\
\midrule
\mbox{0.7, 0.15, 0.15}   & 0.001 & 3.68130 & Tuned \\
\mbox{0.75, 0.125, 0.125}   & 0.001 & 3.48265 & Tuned \\
\mbox{0.8, 0.1, 0.1}   & 0.001 & 3.39890 & Tuned \\
\mbox{\textbf{0.85, 0.075, 0.075}}   & \textbf{0.001} & \textbf{3.38515} & \textbf{Tuned} \\
\mbox{0.85, 0.1, 0.05}   & 0.001 & 3.40070 & Tuned \\
\mbox{0.85, 0.05, 0.1}   & 0.001 & 3.41555 & Tuned \\
\mbox{0.9, 0.05, 0.05}   & 0.001 & 3.44425 & Tuned \\
\bottomrule
\end{tabular}%
}
\caption{Three-source repeat-aware results with a $1/16$ subsample for the 757M model. Mixing ratios are proportions of FineWeb, WikiText, and PubMed. Best run in \textbf{bold}.} 
\label{tab:mixing_results_16_757M}
\end{table*}

\begin{table*}[!t]
\centering
\renewcommand\arraystretch{1.025}
\setlength{\tabcolsep}{25pt}
\scalebox{0.85}{%
\begin{tabular}{l c c c}
\toprule
\makecell{ \textbf{Mixing} \textbf{Ratio}} &
\makecell{ \textbf{Learning} \textbf{Rate}} &
\makecell{ \textbf{Avg.} \textbf{Validation} \textbf{Loss}} &
\makecell{\textbf{Experiment} \textbf{Type}} \\
\midrule
\mbox{0.5, 0.25, 0.25}   & 0.001 & 4.49960 & Tuned \\
\mbox{0.6, 0.2, 0.2}   & 0.001 & 3.86945 & Tuned \\
\mbox{0.7, 0.15, 0.15}   & 0.001 & 3.37725 & Tuned \\
\mbox{\textbf{0.8, 0.1, 0.1}}   & \textbf{0.001} & \textbf{3.20075} & \textbf{Tuned} \\
\mbox{0.8, 0.15, 0.05}   & 0.001 & 3.30435 & Tuned \\
\mbox{0.8, 0.05, 0.15}   & 0.001 & 3.31390 & Tuned \\
\mbox{0.85, 0.075, 0.075}   & 0.001 & 3.20810 & Tuned \\
\mbox{0.9, 0.05, 0.05}   & 0.001 & 3.27125 & Tuned \\
\bottomrule
\end{tabular}%
}
\caption{Three-source repeat-aware results with a $1/8$ subsample for the 757M model. Mixing ratios are proportions of FineWeb, WikiText, and PubMed. Best run in \textbf{bold}.}
\label{tab:mixing_results_8_757M}
\end{table*}

\begin{table*}[!t]
\centering
\renewcommand\arraystretch{1.025}
\setlength{\tabcolsep}{25pt}
\scalebox{0.85}{%
\begin{tabular}{l c c c}
\toprule
\makecell{ \textbf{Mixing} \textbf{Ratio}} &
\makecell{ \textbf{Learning} \textbf{Rate}} &
\makecell{ \textbf{Avg.} \textbf{Validation} \textbf{Loss}} &
\makecell{\textbf{Experiment} \textbf{Type}} \\
\midrule
\mbox{0.7, 0.15, 0.15}   & 0.001 & 3.09955 & Tuned \\
\mbox{0.75, 0.125, 0.125}   & 0.001 & 3.04690 & Tuned \\
\mbox{\textbf{0.8, 0.1, 0.1}}   & \textbf{0.001} & \textbf{3.03955} & \textbf{Tuned} \\
\mbox{0.8, 0.15, 0.05}   & 0.001 & 3.09650 & Tuned \\
\mbox{0.8, 0.05, 0.15}   & 0.001 & 3.11155 & Tuned \\
\mbox{0.85, 0.075, 0.075}   & 0.001 & 3.06200 & Tuned \\
\bottomrule
\end{tabular}%
}
\caption{Three-source repeat-aware results with a $1/4$ subsample for the 757M model. Mixing ratios are proportions of FineWeb, WikiText, and PubMed. Best run in \textbf{bold}.}
\label{tab:mixing_results_4_757M}
\end{table*}

\begin{table*}[!t]
\centering
\renewcommand\arraystretch{1.025}
\setlength{\tabcolsep}{25pt}
\scalebox{0.85}{%
\begin{tabular}{l c c c}
\toprule
\makecell{ \textbf{Mixing} \textbf{Ratio}} &
\makecell{ \textbf{Learning} \textbf{Rate}} &
\makecell{ \textbf{Avg.} \textbf{Validation} \textbf{Loss}} &
\makecell{\textbf{Experiment} \textbf{Type}} \\
\midrule
\mbox{0.65, 0.175, 0.175}   & 0.001 & 2.93330 & Tuned \\
\mbox{0.7, 0.15, 0.15}   & 0.001 & 2.90015 & Tuned \\
\mbox{\textbf{0.75, 0.125, 0.125}}   & \textbf{0.001} & \textbf{2.89195} & \textbf{Tuned} \\
\mbox{0.75, 0.175, 0.075}   & 0.001 & 2.93055 & Tuned \\
\mbox{0.75, 0.075, 0.175}   & 0.001 & 2.94015 & Tuned \\
\mbox{0.8, 0.1, 0.1}   & 0.001 & 2.90955 & Tuned \\
\bottomrule
\end{tabular}%
}
\caption{Three-source repeat-aware results with a $1/2$ subsample for the 757M model. Mixing ratios are proportions of FineWeb, WikiText, and PubMed. Best run in \textbf{bold}.}
\label{tab:mixing_results_2_757M}
\end{table*}

\section{Bootstrap Confidence Intervals}
\label{sec:appendix-bootstrap}
We present the paired 95\% bootstrap confidence intervals for the validation loss of the mixing ratio experiments at the target horizons for the WikiText and PubMed two-source experiments in Table~\ref{tab:bootstrap-ci-1B}. These bootstrap confidence intervals were calculated using 10,000 re-samples.

\section{Multi-Seed Analysis}\label{sec:appendix-seed-variance}
To corroborate the sweep-granularity convention we adopt throughout the paper, we rerun five PubMed configurations under each method (scaling laws and repeat-aware) at three seeds each (42, 43, 44) at the 1.17B scale and measure the resulting spread in validation loss. Table~\ref{tab:seed-variance} reports the results.

Across the ten configurations, validation loss varied by at most 0.0035 for scaling laws and 0.0085 for repeat-aware. This is well below the differences underlying our main claims (e.g., the single-horizon 0.10–0.15 vs. 0.70–0.85 error gaps at 1.17B), and comparable to or smaller than the pairwise gaps in Table~\ref{tab:bootstrap-ci-1B} that we report as significant. The sweep-granularity convention we apply throughout the paper is therefore well-calibrated to the actual run-to-run variability of our training setup.


\begin{table*}[!t]
\centering
\renewcommand\arraystretch{1.025}
\setlength{\tabcolsep}{25pt}
\scalebox{0.85}{%
\begin{tabular}{@{}c c c c c c@{}}
\toprule
\makecell{\textbf{High-Quality}\\\textbf{Dataset}} & \makecell{\textbf{Mixing}\\\textbf{Ratio}} & \makecell{\textbf{Val.}\\\textbf{Loss}} & \makecell{\textbf{Diff vs.}\\\textbf{Next Ratio}} & \makecell{\textbf{95\% CI}\\\textbf{of Diff}} & \makecell{\textbf{Sig.}} \\
\midrule
\multirow{7}{*}{WikiText}
  & 0.60 & 3.6058 & 0.1920 & [0.1783, 0.2059] & Yes \\
  & 0.65 & 3.4138 & 0.2361 & [0.2245, 0.2474] & Yes \\
  & 0.70 & 3.1777 & -0.1221 & [-0.1348, -0.1101] & Yes \\
  & 0.75 & 3.2998 & 0.1853 & [0.1236, 0.2470] & Yes \\
  & 0.80 & 3.1145 & 0.0046 & [-0.0047, 0.0137] & No \\
  & \textbf{0.85} & \textbf{3.1098} & -0.0810 & [-0.0891, -0.0731] & Yes \\
  & 0.90 & 3.1908 & -- & -- & -- \\
\cmidrule(l){2-6}
\multirow{11}{*}{PubMed}
  & 0.40 & 4.4201 & 0.5606 & [0.5339, 0.5869] & Yes \\
  & 0.45 & 3.8595 & 0.3055 & [0.2891, 0.3223] & Yes \\
  & 0.50 & 3.5539 & 0.1255 & [0.1123, 0.1390] & Yes \\
  & 0.55 & 3.4284 & 0.3835 & [0.3701, 0.3964] & Yes \\
  & 0.60 & 3.0449 & 0.1561 & [0.1465, 0.1660] & Yes \\
  & 0.65 & 2.8889 & 0.1549 & [0.1464, 0.1635] & Yes \\
  & 0.70 & 2.7339 & -0.1752 & [-0.1837, -0.1668] & Yes \\
  & 0.75 & 2.9091 & 0.1950 & [0.1879, 0.2024] & Yes \\
  & \textbf{0.80} & \textbf{2.7141} & -0.0096 & [-0.0157, -0.0037] & Yes \\
  & 0.85 & 2.7237 & -0.1447 & [-0.1976, -0.0918] & Yes \\
  & 0.90 & 2.8684 & -- & -- & -- \\
\bottomrule
\end{tabular}%
}
\caption{Validation loss and paired 95\% confidence intervals on the loss difference between each mixing ratio and the next-highest ratio, for the 1.17B target-horizon mixture sweeps (WikiText: 3.74B tokens; PubMed: 3.84B tokens). Mixing ratio is the proportion of FineWeb. Differences and confidence intervals are computed via paired percentile bootstrap (10,000 resamples) over per-sequence validation losses, using the same resampled document indices for both mixtures in each pair so that shared per-document difficulty cancels rather than compounds. The ``Sig.'' column indicates whether the interval excludes zero.}
\label{tab:bootstrap-ci-1B}
\end{table*}

\end{document}